\documentclass{article}

\PassOptionsToPackage{numbers}{natbib}

\usepackage[final]{nips_2018}

\usepackage{natbib}
\usepackage{amsmath}
\usepackage{amssymb}
\usepackage{mathtools}

\usepackage{algorithm}
\usepackage[noend]{algpseudocode}

\usepackage{color}

\usepackage[utf8]{inputenc} 
\usepackage[T1]{fontenc}    
\usepackage{hyperref}       
\usepackage{url}            
\usepackage{booktabs}       
\usepackage{amsfonts}       
\usepackage{microtype}      

\usepackage{graphicx}
\usepackage{subcaption}

\usepackage{appendix}

\usepackage{csvsimple}
\newcommand{\Exp}{\operatorname{\mathbf{E}}}
\newcommand{\R}{\mathbb{R}}

\newcommand{\inv}{^{\raisebox{.2ex}{$\scriptscriptstyle-1$}}}

\newcommand{\covmat}{\mathbf{\Sigma}}
\newcommand{\diag}{\text{diag}}

\DeclareMathOperator*{\argmax}{argmax}

\newcommand{\norm}[1]{\left\|#1\right\|}

\newcommand*\Let[2]{\State #1 $\gets$ #2} 
\newcommand{\ra}[1]{\renewcommand{\arraystretch}{#1}}

\title{Hamiltonian Variational Auto-Encoder}

\author{
  Anthony~L.~Caterini\textsuperscript{1}, Arnaud~Doucet\textsuperscript{1,2}, Dino~Sejdinovic\textsuperscript{1,2} \\
  \textsuperscript{1}Department of Statistics, University of Oxford \\
  \textsuperscript{2}Alan Turing Institute for Data Science\\
  \texttt{\{anthony.caterini, doucet, dino.sejdinovic\}@stats.ox.ac.uk}
}

\begin{document}

\maketitle

\begin{abstract}
Variational Auto-Encoders (VAEs) have become very popular techniques to perform
inference and learning in latent variable models: they allow us
to leverage the rich representational power of neural networks to
obtain flexible approximations of the posterior of latent variables
as well as tight evidence lower bounds (ELBOs). Combined with stochastic
variational inference, this provides a methodology scaling to large datasets.
However, for this methodology to be practically efficient, it is necessary
to obtain low-variance unbiased estimators of the ELBO and its gradients
with respect to the parameters of interest. While the use of Markov chain Monte
Carlo (MCMC) techniques such as Hamiltonian Monte Carlo (HMC) has
been previously suggested to achieve this \cite{salimans2015markov, wolf2016variational}, the proposed methods require
specifying reverse kernels which have a large impact on performance.
Additionally, the resulting unbiased estimator of the ELBO for most
MCMC kernels is typically not amenable to the reparameterization trick.
We show here how to optimally select reverse kernels in this setting
and, by building upon Hamiltonian Importance Sampling (HIS) \cite{neal2005hamiltonian},
we obtain a scheme that provides low-variance unbiased estimators
of the ELBO and its gradients using the reparameterization trick. This allows us to develop a Hamiltonian Variational Auto-Encoder (HVAE). This
method can be re-interpreted as a target-informed normalizing flow \cite{rezende2015variational} which,
within our context, only requires a few evaluations of the gradient
of the sampled likelihood and trivial Jacobian calculations
at each iteration.
\end{abstract}

\section{Introduction}
Variational Auto-Encoders (VAEs), introduced by \citet{kingma2013auto} and \citet{rezende2014stochastic}, are popular techniques to carry out inference and learning in complex latent variable models.
However, the standard mean-field parametrization of the approximate posterior distribution can limit its flexibility.
Recent work has sought to \emph{augment} the VAE approach by sampling from the VAE posterior approximation and transforming these samples through mappings with additional trainable parameters to achieve richer posterior approximations.
The most popular application of this idea is the Normalizing Flows (NFs) approach \cite{rezende2015variational} in which the samples are deterministically evolved through a series of parameterized invertible transformations called a \emph{flow}.
NFs have demonstrated success in various domains \cite{berg2018sylvester, kingma2016improved}, but the flows do not explicitly use information about the target posterior. Therefore, it is unclear whether the improved performance is caused by an accurate posterior approximation or simply a result of overparametrization.
The related Hamiltonian Variational Inference (HVI) \cite{salimans2015markov} instead stochastically evolves the base samples according to Hamiltonian Monte Carlo (HMC) \cite{neal2011mcmc} and thus uses target information, but relies on defining reverse dynamics in the flow, which, as we will see, turns out to be unnecessary and suboptimal.

One of the key components in the formulation of VAEs is the maximization of the evidence lower bound (ELBO).
The main idea put forward in \citet{salimans2015markov} is that it is possible to
use $K$ MCMC iterations to obtain an unbiased estimator of the
ELBO and its gradients.
This estimator is obtained using an importance sampling argument on an augmented space, with the importance
distribution being the joint distribution of the $K+1$ states of the `forward' Markov chain,
while the augmented target distribution is constructed
using a sequence of `reverse' Markov kernels such that it admits the
original posterior distribution as a marginal. The performance of
this estimator is strongly dependent on the selection of these forward and reverse
kernels, but no clear guideline for selection has been provided therein.
By linking this approach to earlier work by \citet{del2006sequential}, we
show how to select these components. We focus, in particular, on the use of time-inhomogeneous
Hamiltonian dynamics, proposed originally in \citet{neal2005hamiltonian}. This method uses reverse kernels which are optimal for reducing variance of the likelihood estimators and allows for simple calculation of the approximate posteriors of the latent variables.
Additionally, we can easily use the reparameterization trick to calculate unbiased gradients of the ELBO with respect to the parameters of interest.
The resulting method, which we refer to as the Hamiltonian Variational Auto-Encoder (HVAE), can be thought of as a Normalizing Flow scheme in which the flow depends explicitly on the target distribution. This combines the best properties of HVI and NFs, resulting in target-informed and inhomogeneous deterministic Hamiltonian dynamics, while being scalable to large datasets and high dimensions.

\section{Evidence Lower Bounds, MCMC and Hamiltonian Importance Sampling}

\subsection{Unbiased likelihood and evidence lower bound estimators}

For data $x\in\mathcal{X}\subseteq\mathbb{R}^{d}$ and parameter $\theta\in\Theta$,
consider the likelihood function
\[
p_{\theta}(x)=\int p_{\theta}(x,z)dz=\int p_{\theta}(x|z)p_{\theta}(z)dz,
\]
where $z\in\mathcal{Z}$ are some latent variables.
If we assume we have
access to a strictly positive unbiased estimate of $p_{\theta}(x)$, denoted $\hat{p}_{\theta}(x)$, then
\begin{equation}
\int\hat{p}_{\theta}(x)q_{\theta,\phi}(u|x)du=p_{\theta}(x),\label{eq:unbiased}
\end{equation}
with $u\sim q_{\theta,\phi}(\cdot)$, $u\in\mathcal{U}$ denoting all the random variables used to compute $\hat{p}_{\theta}(x)$.
Here, $\phi$ denotes additional parameters of the sampling distribution.
We emphasize that $\hat{p}_{\theta}(x)$ depends on both
$u$ and potentially $\phi$ as this is not done notationally. By applying Jensen's
inequality to (\ref{eq:unbiased}), we thus obtain, for all $\theta$ and $\phi$,
\begin{equation} \label{eqn:ELBO_single}
\mathcal{L_{\mathrm{ELBO}}}(\theta,\phi; x)\coloneqq\int\log\hat{p}_{\theta}(x)~q_{\theta, \phi}(u|x)du\leq\log p_{\theta}(x)\eqqcolon\mathcal{L}(\theta; x).
\end{equation}
It can be shown that $|\mathcal{L}_{\mathrm{ELBO}}(\theta,\phi; x)-\mathcal{L}(\theta; x)|$
decreases as the variance of $\hat{p}_{\theta}(x)$ decreases;
see, e.g., \cite{burda2015importance, maddison2017filtering}. The standard variational
framework corresponds to $\mathcal{U}=\mathcal{Z}$ and $\hat{p}_{\theta}(x)=p_{\theta}(x,z)/q_{\theta,\phi}(z|x)$, while the Importance Weighted Auto-Encoder (IWAE) \cite{burda2015importance} with $L$ importance samples corresponds to $\mathcal{U}=\mathcal{Z}^{L}$, $q_{\theta,\phi}(u|x)=\prod_{i=1}^{L}q_{\theta,\phi}(z_{i}|x)$
and $\hat{p}_{\theta}(x)=\frac{1}{L}\sum_{i=1}^{L}p_{\theta}(x,z_{i})/q_{\theta,\phi}(z_{i}|x).$

In the general case, we do not have an analytical expression for $\mathcal{L}_{\mathrm{ELBO}}(\theta,\phi; x)$.
When performing stochastic gradient ascent for variational inference, however, we only require an unbiased estimator of $\nabla_{\theta}\mathcal{L}_{\mathrm{ELBO}}(\theta, \phi; x)$.
This is given by $\nabla_{\theta}\log\hat{p}_{\theta}(x)$
if the reparameterization trick \cite{glasserman1991gradient,kingma2013auto} is used, i.e. $q_{\theta,\phi}(u|x)=q(u)$, and $\hat{p}_{\theta}(x)$ is a `smooth enough' function
of $u$. As a guiding principle, one should attempt to obtain a low-variance estimator of $p_{\theta}(x)$, which typically
translates into a low-variance estimator of $\nabla_\theta \mathcal{L}_{\mathrm{ELBO}}(\theta, \phi; x)$.
We can analogously optimize $\mathcal{L_{\mathrm{ELBO}}}(\theta,\phi; x)$ with respect to $\phi$
through stochastic gradient ascent to obtain tighter bounds.

\subsection{Unbiased likelihood estimator using time-inhomogeneous MCMC}

\citet{salimans2015markov} propose to build an unbiased
estimator of $p_{\theta}(x)$ by sampling a (potentially time-inhomogeneous) `forward' Markov
chain of length $K+1$ using $z_{0}\sim q_{\theta,\phi}^{0}(\cdot)$
and $z_{k}\sim q_{\theta,\phi}^{k}(\cdot|z_{k-1})$ for $k=1,...,K$.
Using artificial `reverse' Markov transition kernels $r_{\theta,\phi}^{k}(z_{k}|z_{k+1})$
for $k=0,...,K-1$, it follows easily from an importance sampling
argument that
\begin{equation}
\hat{p}_{\theta}(x)=\frac{p_{\theta}(x,z_{K})\prod_{k=0}^{K-1}r_{\theta,\phi}^{k}(z_{k}|z_{k+1})}{q_{\theta,\phi}^{0}(z_{0})\prod_{k=1}^{K}q_{\theta,\phi}^{k}(z_{k}|z_{k-1})}\label{eq:generalizedJarzynski}
\end{equation}
is an unbiased estimator of $p_{\theta}(x)$ as long
as the ratio in (\ref{eq:generalizedJarzynski}) is well-defined.
In the framework of the previous section, we have $\mathcal{U}=\mathcal{Z}^{K+1}$
and $q_{\theta,\phi}(u|x)$ is given by the denominator
of (\ref{eq:generalizedJarzynski}). Although we did not use measure-theoretic
notation, the kernels $q_{\theta,\phi}^{k}$ are typically MCMC kernels
which do not admit a density with respect to the Lebesgue measure (e.g.
the Metropolis\textendash Hastings kernel). This makes it difficult
to define reverse kernels for which (\ref{eq:generalizedJarzynski})
is well-defined as evidenced in \citet[Section 4]{salimans2015markov}
or \citet{wolf2016variational}. The estimator (\ref{eq:generalizedJarzynski}) was originally introduced in \citet{del2006sequential} where generic recommendations are provided for this estimator to admit a low relative variance: select $q_{\theta,\phi}^{k}$ as MCMC kernels which are
invariant, or approximately invariant as in \cite{heng2017controlled}, with respect to $p^k_\theta(x,z_{k})$, where $p_{\theta,\phi}^{k}(z|x)\propto \left[q_{\theta,\phi}^{0}(z)\right]^{1-\beta_{k}} \left[p_{\theta}(x,z)\right]^{\beta_{k}}$
is a sequence of artificial densities bridging $q_{\theta,\phi}^{0}(z)$
to $p_{\theta}(z|x)$ smoothly using $\beta_{0}=0<\beta_{1}<\cdots<\beta_{K-1}<\beta_{K}=1$.
It is also established in \citet{del2006sequential} that, given any sequence of kernels $\{q_{\theta,\phi}^{k}\}_k$,
the sequence of reverse kernels minimizing the variance of $\hat{p}_{\theta}(x)$
is given by $r_{\theta,\phi}^{k,\mathrm{opt}}(z_{k}|z_{k+1})=q_{\theta,\phi}^{k}(z_{k})q_{\theta,\phi}^{k+1}(z_{k+1}|z_{k})/q_{\theta,\phi}^{k+1}(z_{k+1}), $
where $q_{\theta,\phi}^{k}\left(z_{k}\right)$ denotes the marginal
density of $z_{k}$ under the forward dynamics, yielding
\begin{equation}
\hat{p}_{\theta}(x)=\frac{p_{\theta}(x,z_{K})}{q_{\theta,\phi}^{K}(z_{K})}.\label{eq:optimalestimator}
\end{equation}
For stochastic forward transitions, it is typically not possible to compute $r_{\theta,\phi}^{k,\mathrm{opt}}$ and the corresponding estimator (\ref{eq:optimalestimator}) as the marginal densities $q_{\theta,\phi}^{k}(z_{k})$ do not admit closed-form expressions.
However this suggests that  $r_{\theta,\phi}^{k}$ should be approximating $r_{\theta,\phi}^{k,\mathrm{opt}}$ and various schemes are presented in \cite{del2006sequential}.
As noticed by \citet{del2006sequential} and \citet{salimans2015markov}, Annealed Importance Sampling (AIS)
\cite{neal2001annealed} -- also known as the Jarzynski-Crooks identity (\cite{crooks1998nonequilibrium, jarzynski1997nonequilibrium}) in physics -- is a special case of (\ref{eq:generalizedJarzynski})
using, for $q_{\theta,\phi}^{k}$, a $p_{\theta}^{k}(z|x)$-invariant
MCMC kernel and the reversal of this kernel as the reverse transition kernel $r_{\theta,\phi}^{k-1}$\footnote{The reversal of a $\mu$-invariant kernel $K(z'|z)$ is given by $K_{rev}(z'|z)=\frac{\mu(z')K(z|z')}{\mu(z)}$. If $K$ is $\mu$-reversible then $K_{rev}=K$.}. This choice of reverse kernels is suboptimal but leads to a simple expression for estimator  (\ref{eq:generalizedJarzynski}).
AIS provides state-of-the-art estimators of the marginal likelihood and has been widely used in machine learning.
Unfortunately, it typically cannot be used in conjunction
with the reparameterization trick. Indeed, although it is very often
possible to reparameterize the forward simulation of $(z_{1},...,z_{T})$
in terms of the deterministic transformation of some random variables
$u\sim q$ independent of $\theta$ and $\phi$, this mapping is not continuous
because the MCMC kernels it uses typically include singular components. In this context, although
(\ref{eq:unbiased}) holds, $\nabla_\theta\log\hat{p}_{\theta}(x)$
\emph{is not an unbiased estimator of} $\nabla_\theta\mathcal{L}_{\mathrm{ELBO}}(\theta, \phi; x)$; see, e.g., \citet{glasserman1991gradient} for a careful
discussion of these issues.

\subsection{Using Hamiltonian dynamics}\label{sec:deterministic}

Given the empirical success of Hamiltonian Monte Carlo (HMC) \cite{hoffman2014no, neal2011mcmc}, various contributions have proposed to develop algorithms
exploiting Hamiltonian dynamics to obtain unbiased estimates of the ELBO and its gradients when $\mathcal{Z}=\mathbb{R}^\ell$. This was proposed in \citet{salimans2015markov}. However, the algorithm suggested therein relies on a time-homogeneous leapfrog where momentum resampling is performed at each step and no Metropolis correction is used.
It also relies on learned reverse kernels.
To address the limitations of \citet{salimans2015markov}, \citet{wolf2016variational} have proposed to include some Metropolis acceptance steps, but they still limit themselves to homogeneous dynamics and their estimator is not amenable to the reparameterization trick. Finally, in \citet{hoffman2017learning}, an alternative approach is used where the gradient of the true likelihood, $\nabla_\theta\mathcal{L}(\theta; x)$, is directly approximated
by using Fisher's identity and HMC to obtain approximate samples from $p_{\theta}(z|x)$.
However, the MCMC bias can be very significant when one has multimodal latent posteriors and is strongly dependent on both the initial distribution and $\theta$.

Here, we follow an alternative approach where we use Hamiltonian dynamics that are time-inhomogeneous as in \cite{del2006sequential} and \cite{neal2001annealed}, and use optimal reverse Markov kernels to compute $\hat{p}_{\theta}(x)$. This estimator can be used in conjunction with the reparameterization trick to obtain
an unbiased estimator of $\nabla\mathcal{L}_{\mathrm{ELBO}}(\theta, \phi; x)$.
This method is based on the Hamiltonian Importance Sampling (HIS) scheme proposed
in \citet{neal2005hamiltonian}; one can also find several instances of related
ideas in physics \cite{jarzynski2000hamiltonian,scholl2006proof}. We work in an extended space $(z,\rho)\in \cal{U}:=\mathbb{R}^\ell\times\mathbb{R}^\ell$, introducing \emph{momentum} variables $\rho$ to pair with the \emph{position} variables $z$,
with new target $\bar{p}_{\theta}(x,z,\rho):=p_{\theta}(x,z)\mathcal{N}(\rho|0, I_\ell)$.
Essentially, the idea is to sample using \emph{deterministic} transitions $q_{\theta,\phi}^{k}((z_{k},\rho_{k})|(z_{k-1},\rho_{k-1}))=\delta_{\Phi_{\theta,\phi}^{k}(z_{k-1},\rho_{k-1})}(z_{k},\rho_{k})$
so that $(z_{K},\rho_{K})=\mathcal{H}_{\theta,\phi}(z_{0},\rho_{0}):=\left(\Phi_{\theta,\phi}^{K}\circ\cdots\circ\Phi_{\theta, \phi}^{1}\right)\left(z_{0},\rho_{0}\right)$,
where $(z_{0},\rho_{0})\sim q_{\theta,\phi}^{0}(\cdot,\cdot)$ and
$(\Phi_{\theta,\phi}^{k})_{k\geq1}$, define diffeomorphisms corresponding
to a time-discretized and inhomogeneous Hamiltonian dynamics. In this
case, it is easy to show that
\begin{equation}
q_{\theta,\phi}^{K}(z_{K},\rho_{K})=q_{\theta,\phi}^{0}(z_{0},\rho_{0})\prod_{k=1}^{K}\left|\det\nabla\Phi^k_{\theta,\phi}(z_{k},\rho_{k})\right|^{-1}\quad \text{and} \quad\hat{p}_{\theta}(x)=\frac{\bar{p}_{\theta}(x,z_{K},\rho_{K})}{q_{\theta,\phi}^{K}(z_{K},\rho_{K})}.\label{eq:estimatorHMC}
\end{equation}
It can also be shown that this is nothing but a special case of (\ref{eq:generalizedJarzynski})
(on the extended position-momentum space) using the optimal reverse kernels\footnote{Since this is a deterministic flow, the density can be evaluated directly. However, direct evaluation corresponds to optimal reverse kernels in the deterministic case.} $r_{\theta,\phi}^{k,\mathrm{opt}}$.
This setup is similar to the one of Normalizing Flows \cite{rezende2015variational},
except here we use a flow informed by the target distribution.   \citet{salimans2015markov} is in fact mentioned in \citet{rezende2015variational}, but the flow therein
is homogeneous and yields a high-variance estimator of the normalizing
constants even if $r_{\theta}^{k,\mathrm{opt}}$ is used, as demonstrated
in our simulations in \autoref{sec:experiments}.

Under these dynamics, the estimator $\hat{p}_{\theta}(x)$ defined
in (\ref{eq:estimatorHMC}) can be rewritten as
\begin{equation} \label{eqn:reparameterization}
\hat{p}_{\theta}(x)=\frac{\bar{p}_{\theta}\left(x,\mathcal{H}_{\theta,\phi}\left(z_{0},\rho_{0}\right)\right)}{q_{\theta, \phi}^{0}\left(z_{0},\rho_{0}\right)}\prod_{k=1}^{K}\left|\det\nabla\Phi^k_{\theta,\phi}(z_{k},\rho_{k})\right|.
\end{equation}
Hence, if we can simulate $(z_{0},\rho_{0})\sim q_{\theta, \phi}^{0}(\cdot,\cdot)$
using $(z_{0},\rho_{0})=\varPsi_{\theta,\phi}(u)$, where $u\sim q$
and $\varPsi_{\theta,\phi}$ is a smooth mapping, then we can
use the reparameterization trick since $\Phi_{\theta,\phi}^{k}$ are also smooth mappings.

In our case, the deterministic transformation $\Phi^k_{\theta,\phi}$ has two components: a leapfrog step, which discretizes the Hamiltonian dynamics, and a tempering step, which adds inhomogeneity to the dynamics and allows us to explore isolated modes of the target \cite{neal2005hamiltonian}. To describe the leapfrog step, we first define the potential energy of the system as $U_\theta(z | x) \equiv -\log p_\theta(x, z)$ for a single datapoint $x \in \mathcal{X}$. Leapfrog then takes the system from $(z, \rho)$ into $(z', \rho')$ via the following transformations:
\begin{align}
\widetilde \rho &= \rho - \frac{\varepsilon}{2} \odot \nabla U_\theta(z|x), \label{eqn:lf1} \\
z' &= z + \varepsilon \odot \widetilde \rho, \label{eqn:lf2} \\
\rho' &= \widetilde \rho - \frac{\varepsilon}{2} \odot \nabla U_\theta(z' | x), \label{eqn:lf3}
\end{align}
where $\varepsilon \in (\R^{+})^\ell$ are the individual leapfrog step sizes per dimension, $\odot$ denotes elementwise multiplication, and the gradient of $U_\theta(z | x)$ is taken with respect to $z$.
The composition of equations \eqref{eqn:lf1} - \eqref{eqn:lf3} has unit Jacobian since each equation describes a shear transformation.
For the tempering portion, we multiply the momentum output of each leapfrog step by $\alpha_k \in (0,1)$ for $k \in [K]$ where $[K] \equiv \{1,\ldots, K\}$.
We consider two methods for setting the values $\alpha_k$.
First, \emph{fixed tempering} involves allowing an inverse temperature $\beta_0 \in (0,1)$ to vary, and then setting $\alpha_k = \sqrt{\beta_{k-1}/ \beta_k}$, where each $\beta_k$ is a deterministic function of $\beta_0$ and $0 < \beta_0 < \beta_1 < \ldots < \beta_K = 1$.
 In the second method, known as \emph{free tempering}, we allow each of the $\alpha_k$ values to be learned, and then set the initial inverse temperature to $\beta_0 = \prod_{k=1}^K \alpha_k^2$.
For both methods, the tempering operation has Jacobian $\alpha_k^\ell$.
We obtain $\Phi^k_{\theta, \phi}$ by
      composing the leapfrog integrator with the cooling operation, which implies that the Jacobian is given by $|\det \nabla \Phi^k_{\theta, \phi}(z_k, \rho_k) | = \alpha_k^\ell = (\beta_{k-1} / \beta_k)^{\ell/2}$, which in turns implies
\[
\prod_{k=1}^K |\det \nabla \Phi^k_{\theta,\phi} (z_k, \rho_k)| = \prod_{k=1}^K \left(\frac{\beta_{k-1}}{\beta_k}\right)^{\ell/2} = \beta_0^{\ell/2}.
\]

The only remaining component to specify is the initial distribution. We will set $q^0_{\theta, \phi}(z_0, \rho_0) = q^0_{\theta,\phi}(z_0) \cdot \mathcal{N}(\rho_0 |0, \beta_0\inv I_\ell)$, where $q_{\theta,\phi}^0(z_0)$ will be referred to as the \emph{variational prior} over the latent variables and $\mathcal{N}(\rho_0 |0, \beta_0\inv I_\ell)$ is the \emph{canonical momentum distribution} at inverse temperature $\beta_0$.
The full procedure to generate an unbiased estimate of the ELBO from \eqref{eqn:ELBO_single} on the extended space $\mathcal{U}$ for a single point $x \in \mathcal{X}$ and fixed tempering is given in \autoref{alg:his}.
The set of variational parameters to optimize contains the flow parameters $\beta_0$ and $\varepsilon$, along with additional parameters of the variational prior.\footnote{We avoid reference to a mass matrix $M$ throughout this formulation because we can capture the same effect by optimizing individual leapfrog step sizes per dimension as pointed out in \cite[Section 4.2]{neal2011mcmc}.}
We can see from \eqref{eqn:reparameterization} that we will obtain unbiased gradients with respect to $\theta$ and $\phi$ from our estimate of the ELBO if we write $(z_0, \rho_0) = \left(z_0, \gamma_0 / \sqrt{\beta_0}\right)$, for $z_0 \sim q^0_{\theta,\phi}(\cdot)$ and $\gamma_0 \sim \mathcal{N}(\cdot | 0, I_\ell) \equiv \mathcal{N}_\ell(\cdot)$, provided we are not also optimizing with respect to parameters of the variational prior.
We will require additional reparameterization when we elect to optimize with respect to the parameters of the variational prior, but this is generally quite easy to implement on a problem-specific basis and is well-known in the literature; see, e.g. \cite{kingma2013auto, rezende2015variational, rezende2014stochastic} and \autoref{sec:experiments}.

\begin{algorithm}
\caption{Hamiltonian ELBO, Fixed Tempering}
\label{alg:his}
\begin{algorithmic}
	\Require{$p_\theta(x, \cdot)$ is the unnormalized posterior for $x \in \mathcal{X}$ and $\theta \in \Theta$} \Require{$q_{\theta,\phi}^0(\cdot)$ is the variational prior on $\R^\ell$}
	\Statex
	\Function{HIS}{$x, \theta, K, \beta_0, \varepsilon$}
		\State Sample $z_0 \sim q^0_{\theta,\phi}(\cdot), \gamma_0 \sim \mathcal{N}_\ell(\cdot)$
		\Let{$\rho_0$}{$\gamma_0 / \sqrt{\beta_0}$} \Comment $\rho_0 \sim \mathcal{N}(\cdot|0, \beta_0\inv I_\ell)$
		\For{$k \gets 1 \textrm{ to } K$} \Comment Run $K$ steps of alternating leapfrog and tempering
			\Let{$\widetilde \rho$}{$\rho - \varepsilon /{2} \odot  \nabla U_\theta(z_{k-1} | x)$} \Comment Start of leapfrog; Equation \eqref{eqn:lf1}
			\Let{$z_k$}{$z_{k-1} + \varepsilon \odot \widetilde \rho$} \Comment Equation \eqref{eqn:lf2}
			\Let{$\rho'$}{$\widetilde \rho - \varepsilon / {2} \odot \nabla U_\theta(z_k | x)$} \Comment Equation \eqref{eqn:lf3}
			
			\vspace{0.2em}
			\Let{$\sqrt{\beta_k}$}{$\left(\left(1 - \frac{1}{\sqrt{\beta_0}}\right) \cdot k^2 /{K^2} + \frac{1}{\sqrt{\beta_0}}\right)\inv$} \Comment{Quadratic tempering scheme}
			\Let{$\rho_k$}{$\sqrt{\beta_{k-1} / \beta_k} \cdot \rho'$}
		\EndFor
		\Let{$\bar p$}{$p_\theta(x, z_K) \mathcal{N}(\rho_K | 0, I_\ell)$}
		\Let{$\bar q$}{$q_{\theta,\phi}^0(z_0) \mathcal{N}(\rho_0 | 0, \beta_0\inv I_\ell) \beta_0^{-\ell/{2}}$} \Comment Equation \eqref{eq:estimatorHMC}, left side
		\Let{$\hat{\mathcal{L}}^H_{\mathrm{ELBO}}(\theta, \phi; x)$}{$\log \bar p - \log\bar q$} \Comment Take the $\log$ of equation \eqref{eq:estimatorHMC}, right side
		\State \Return $\hat{\mathcal{L}}^H_{\mathrm{ELBO}}(\theta, \phi; x)$ \Comment Can take unbiased gradients of this estimate wrt $\theta, \phi$
	\EndFunction
\end{algorithmic}
\end{algorithm}

\section{Stochastic Variational Inference}
We will now describe how to use \autoref{alg:his} within a stochastic variational inference procedure, moving to the setting where we have a dataset $\mathcal{D} = \{x_1, \ldots, x_N\}$ and $x_i \in \mathcal{X}$ for all $i \in [N]$. In this case, we are interested in finding
\begin{equation} \label{eqn:inference_objective}
\theta^* \in \argmax_{\theta \in \Theta} \Exp_{x \sim \nu_\mathcal{D}(\cdot)}[\mathcal{L}(\theta; x)],
\end{equation}
where $\nu_{\mathcal{D}}(\cdot) \equiv\frac{1}{N} \sum_{i=1}^N\delta_{x_i}(\cdot)$ is the empirical measure of the data.
We must resort to variational methods since $\mathcal{L}(\theta; x)$ cannot generally be calculated exactly and instead maximize the surrogate ELBO objective function
\begin{equation} \label{eqn:ELBO}
\mathcal{L}_{\mathrm{ELBO}}(\theta, \phi) \equiv \Exp_{x \sim \nu_{\mathcal{D}}(\cdot)} \left[ \mathcal{L}_{\mathrm{ELBO}}(\theta, \phi; x) \right]
\end{equation}
for $\mathcal{L}_{\mathrm{ELBO}}(\theta, \phi; x)$ defined as in \eqref{eqn:ELBO_single}. We can now turn to stochastic gradient ascent (or a variant thereof) to jointly maximize \eqref{eqn:ELBO} with respect to $\theta$ and $\phi$ by approximating the expectation over $\nu_\mathcal{D}(\cdot)$ using \emph{minibatches} of observed data.

For our specific problem, we can reduce the variance of the ELBO calculation by analytically evaluating some terms in the expectation (i.e. Rao-Blackwellization) as follows:
\begin{align}
\mathcal{L}^H_{\mathrm{ELBO}}(\theta, \phi; x) &= \Exp_{(z_0, \rho_0) \sim q_{\theta,\phi}^{0}(\cdot, \cdot)}\left[\log \left(\frac{\bar p_\theta(x, z_K, \rho_K)\beta_0^{\ell/{2}}}{q_{\theta, \phi}^{0}(z_0, \rho_0)} \right)\right] \nonumber \\
&= \Exp_{z_0 \sim q_{\theta,\phi}^{0}(\cdot), \gamma_0 \sim \mathcal{N}_\ell(\cdot)}\left[\log p_\theta(x, z_K)\label{eqn:H_ELBO_simplified}
- \frac{1}{2} \rho_K^T \rho_K - \log q_{\theta,\phi}^0(z_0) \right] +\frac{\ell}{2},
\end{align}
where we write $(z_K, \rho_K) = \mathcal{H}_{\theta, \phi}\left(z_0, \gamma_0 / \sqrt{\beta_0}\right)$ under reparameterization. We can now consider the output of \autoref{alg:his} as taking a sample from the inner expectation for a given sample $x$ from the outer expectation.
\autoref{alg:vi} provides a full procedure to stochastically optimize \eqref{eqn:H_ELBO_simplified}.
In practice, we take the gradients of \eqref{eqn:H_ELBO_simplified} using automatic differentation packages. This is achieved by using TensorFlow  \cite{tensorflow2015-whitepaper} in our implementation.

\begin{algorithm}
\caption{Hamiltonian Variational Auto-Encoder}
\label{alg:vi}
\begin{algorithmic}
	\Require{$p_\theta(x, \cdot)$ is the unnormalized posterior for $x \in \mathcal{X}$ and $\theta \in \Theta$}
	\Statex
	\Function{HVAE}{$\mathcal{D}, K, n_B$} \Comment $n_B$ is minibatch size
		\State Initialize $\theta, \phi$
		\While{$\theta, \phi$ not converged} \Comment Stochastic optimization loop
			\State Sample $\{x_1, \ldots, x_{n_B}\} \sim \nu_{\mathcal{D}}(\cdot)$ independently
			\Let{$\hat{\mathcal{L}}^H_{\mathrm{ELBO}}(\theta, \phi)$}{$0$} \Comment Average ELBO estimators over mini-batch
			\For{$i \gets 1 \textrm{ to } n_B$}
				\Let{$\hat{\mathcal{L}}^H_{\mathrm{ELBO}}(\theta, \phi)$}{\Call{HIS}{$x_i, \theta, K, \beta_0, \varepsilon$} + $\hat{\mathcal{L}}^H_{\mathrm{ELBO}}(\theta, \phi)$}
			\EndFor
            \Let{$\hat{\mathcal{L}}^H_{\mathrm{ELBO}}(\theta, \phi)$}{$\hat{\mathcal{L}}^H_{\mathrm{ELBO}}(\theta, \phi)/n_B$}
			\State \Comment \hspace{0.5cm} Optimize the ELBO using gradient-based techniques such as RMSProp, ADAM, etc.
			\Let{$\theta$}{\Call{UpdateTheta}{$\nabla_\theta \hat{\mathcal{L}}^H_{\mathrm{ELBO}}(\theta, \phi), \theta$}}
			\Let{$\phi$}{\Call{UpdatePhi}{$\nabla_\phi \hat{\mathcal{L}}^H_{\mathrm{ELBO}}(\theta, \phi), \phi$}}
		\EndWhile
		\State \Return $\theta, \phi$
	\EndFunction
\end{algorithmic}
\end{algorithm}
\section{Experiments}\label{sec:experiments}

In this section, we discuss the experiments used to validate our method.
We first test HVAE on an example with a tractable full log likelihood (where no neural networks are needed), and then perform larger-scale tests on the MNIST dataset.
Code is available online.\footnote{\url{https://github.com/anthonycaterini/hvae-nips}} All models were trained using TensorFlow \cite{tensorflow2015-whitepaper}.

\subsection{Gaussian Model}
The generative model that we will consider first is a Gaussian likelihood with an offset and a Gaussian prior on the mean, given by
\begin{align*}
z &\sim \mathcal{N}(0, I_\ell), \\
x_i | z &\sim \mathcal{N}(z + \Delta, \covmat) \quad \text{independently}, \qquad i \in [N]
\end{align*}
where $\covmat$ is constrained to be diagonal. We will again write $\mathcal{D} \equiv \{x_1, \ldots, x_N\}$ to denote an observed dataset under this model, where each $x_i \in \mathcal{X} \subseteq \R^d$.
In this example, we have $\ell = d$.
The goal of the problem is to learn the model parameters $\theta \equiv \{\covmat, \Delta\}$, where $\covmat = \diag(\sigma_1^2, \ldots, \sigma_d^2)$ and $\Delta \in \R^d$.

Here, we have only one latent variable generating the entire set of data. Thus, our variational lower bound is now given by
\[
\mathcal{L}_{\mathrm{ELBO}}(\theta, \phi; \mathcal{D}) \coloneqq \Exp_{z \sim q_{\theta, \phi}(\cdot | \mathcal{D})} \left[ \log p_\theta(\mathcal{D}, z) - \log q_{\theta, \phi}(z | \mathcal{D} )\right] \leq \log p_\theta(\mathcal{D}),
\]
for the variational posteroir approximation $q_{\theta, \phi}(\cdot | \mathcal{D})$. We note that this is not exactly the same as the auto-encoder setting, in which an individual latent variable is associated with each observation, however it provides a tractable framework to analyze effectiveness of various variational inference methods. We also note that we can calculate the log-likelihood $\log p_\theta(\mathcal{D})$ exactly in this case, but we use variational methods for the sake of comparison.

From the model, we see that the logarithm of the unnormalized target is given by
\[
\log p_\theta(\mathcal{D}, z) = \sum_{i=1}^N \log \mathcal{N}(x_i | z + \Delta, \covmat) + \log \mathcal{N}(z | 0, I_d). 
\]

For this example, we will use a HVAE with variational prior equal to the true prior, i.e. $q^0 = \mathcal{N}(0,I_\ell)$, and fixed tempering. The potential, given by $U_\theta(z | \mathcal{D}) = \log p_\theta(\mathcal{D}, z)$, has gradient
\[
\nabla U_\theta(z | \mathcal{D}) = z + N \covmat\inv(z + \Delta - x).
\]
The set of variational parameters here is $\phi \equiv \{\varepsilon, \beta_0\}$, where $\varepsilon \in \R^d$ contains the per-dimension leapfrog stepsizes and $\beta_0 \in (0,1)$ is the initial inverse temperature. We constrain each of the leapfrog step sizes such that $\varepsilon_j \in (0, \xi)$ for some $\xi > 0$, for all $j \in [d]$ -- this is to prevent the leapfrog discretization from entering unstable regimes. Note that $\phi \in \R^{d+1}$ in this example; in particular, we do not optimize any parameters of the variational prior and thus require no further reparameterization.

We will compare HVAE with a basic Variational Bayes (VB) scheme with mean-field approximate posterior $q_{\phi_V} (z | \mathcal{D}) = \mathcal{N}(z | \mu_Z, \covmat_Z)$, where $\covmat_Z$ is diagonal and $\phi_V \equiv \{\mu_Z, \covmat_Z\}$ denotes the set of learned variational parameters.
We will also include a planar normalizing flow of the form of equation (10) in \citet{rezende2015variational}, but with the same flow parameters across iterations to keep the number of variational parameters of the same order as the other methods.
The variational prior here is also set to the true prior as in HVAE above.
The log variational posterior $\log q_{\phi_N}(z | \mathcal{D})$ is given by equation (13) of \citet{rezende2015variational}, where $\phi_N \equiv \{\mathbf{u}, \mathbf{v}, b\}\footnote{Boldface vectors used to match notation of \citet{rezende2015variational}.} \in \R^{2d+1}$.

We set our true offset vector to be $\Delta = \left(- \frac{d-1}{2}, \ldots, \frac{d-1}{2}\right) / 5$, and our scale parameters to range quadratically from $\sigma_1 = 1$, reaching a minimum at $\sigma_{(d+1)/2} = 0.1$,  and increasing back to $\sigma_d = 1$.\footnote{When $d$ is even, $\sigma_{(d+1)/2}$ does not exist, although we could still consider $(d+1)/2$ to be the location of the minimum of the parabola defining the true standard deviations.} All experiments have $N = 10,\!000$ and all training was done using RMSProp \cite{tieleman2012lecture} with a learning rate of $10^{-3}$.

\begin{figure}
	\centering
	\begin{subfigure}[b]{0.49\textwidth}
		\includegraphics[width=\textwidth]{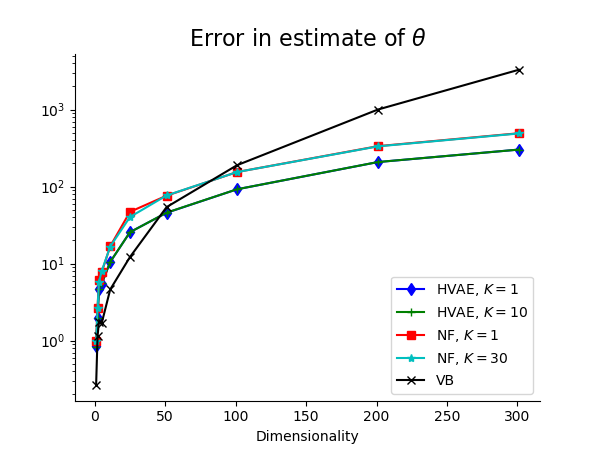}
  	\caption{Comparison across all methods}
  	\label{fig:all_methods}
  \end{subfigure}
	\begin{subfigure}[b]{0.49\textwidth}
		\includegraphics[width=\textwidth]{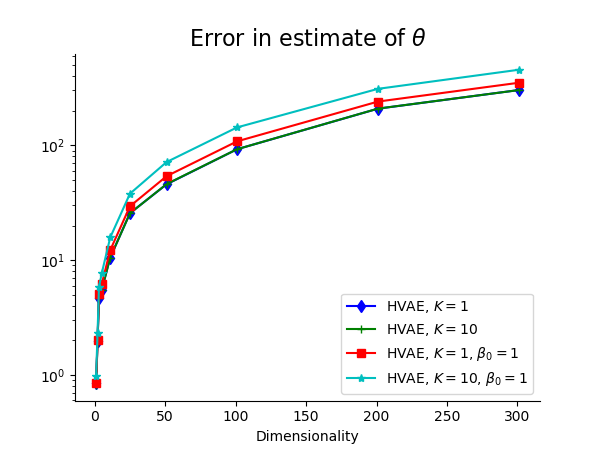}
  	\caption{HVAE with and without tempering}
  	\label{fig:temp_vs_no_temp}
  \end{subfigure}
  \caption{Averages of $\norm{\theta - \hat \theta}_2^2$ for several variational methods and choices of dimensionality $d$, where $\hat \theta$ is the estimated maximizer of the ELBO for each method and $\theta$ is the true parameter.}
\end{figure}

To compare the results across methods, we train each method ten times on different datasets. For each training run, we calculate $\norm{\theta - \hat \theta}^2_2$, where $\hat \theta$ is the estimated value of $\theta$ given by the variational method on a particular run, and plot the average of this across the 10 runs for various dimensions in \autoref{fig:all_methods}. We note that, as the dimension increases, HVAE performs best in parameter estimation. The VB method suffers most on prediction of $\Delta$ as the dimension increases, whereas the NF method does poorly on predicting $\covmat$.

We also compare HVAE with tempering to HVAE without tempering, i.e. where $\beta_0$ is fixed to $1$ in training. This has the effect of making our Hamiltonian dynamics homogeneous in time. We perform the same comparison as above and present the results in \autoref{fig:temp_vs_no_temp}. We can see that the tempered methods perform better than their non-tempered counterparts; this shows that time-inhomogeneous dynamics are a key ingredient in the effectiveness of the method.

\subsection{Generative Model for MNIST}

The next experiment that we consider is using HVAE to improve upon a convolutional variational auto-encoder (VAE) for the binarized MNIST handwritten digit dataset.
Again, our training data is $\mathcal{D} = \{x_1, \ldots, x_N\}$, where each $x_i \in \mathcal{X} \subseteq \{0,1\}^d$ for $d = 28 \times 28 = 784$.
The generative model is as follows:
\begin{align*}
z_i &\sim \mathcal{N}(0, I_\ell), \\
x_i | z_i &\sim \prod_{j=1}^d \text{Bernoulli}((x_i)_j | \pi_\theta(z_i)_j),
\end{align*}
for $i\in [N]$, where $(x_i)_j$ is the $j^{th}$ component of $x_i$, $z_i \in \mathcal{Z} \equiv \R^\ell$ is the latent variable associated with $x_i$, and $\pi_\theta : \mathcal{Z} \rightarrow \mathcal{X}$ is a convolutional neural network (i.e. the \emph{generative network}, or \emph{encoder}) parametrized by the model parameters $\theta$.
This is the standard generative model used in VAEs in which each pixel in the image $x_i$ is conditionally independent given the latent variable.
The VAE approximate posterior -- and the HVAE \emph{variational prior} across the latent variables in this case -- is given by $q_{\theta,\phi}(z_i | x_i) = \mathcal{N}\left(z_i | \mu_{\phi}(x_i), \covmat_{\phi}(x_i)\right)$, where $\mu_{\phi}$ and $\covmat_{\phi}$ are separate outputs of the same neural network (the \emph{inference network}, or \emph{encoder}) parametrized by $\phi$, and $\covmat_{\phi}$ is constrained to be diagonal.

We attempted to match the network structure of \citet{salimans2015markov}. The inference network consists of three convolutional layers, each with filters of size $5 \times 5$ and a stride of 2. The convolutional layers output 16, 32, and 32 feature maps, respectively. The output of the third layer is fed into a fully-connected layer with hidden dimension $n_h = 450$, whose output is then fully connected to the output means and standard deviations each of size $\ell$. Softplus activation functions are used throughout the network except immediately before the outputted mean. The generative network mirrors this structure in reverse, replacing the stride with upsampling as in \citet{dosovitskiy2015learning} and replicated in \citet{salimans2015markov}.

We apply HVAE on top of the base convolutional VAE. We evolve samples from the variational prior according to \autoref{alg:his} and optimize the new objective given in \eqref{eqn:H_ELBO_simplified}.
We reparameterize $z_0 | x \sim \mathcal{N}\left(\mu_{\phi}(x), \covmat_{\phi}(x)\right)$ as $z_0 = \mu_{\phi}(x) + \covmat^{1/{2}}_{\phi}(x)\cdot \epsilon$, for $\epsilon \sim \mathcal{N}(0, I_\ell)$ and $x \in \mathcal{X}$, to generate unbiased gradients of the ELBO with respect to $\phi$.
We select various values for $K$ and set $\ell = 64$.
In contrast with normalizing flows, we do not need our flow parameters $\varepsilon$ and $\beta_0$ to be outputs of the inference network because our flow is guided by the target.
This allows our method to have fewer overall parameters than normalizing flow schemes.
We use the standard stochastic binarization of MNIST \cite{salakhutdinov2008quantitative} as training data, and train using Adamax \cite{kingma2014adam} with learning rate $10^{-3}$.
We also employ early stopping by halting the training procedure if there is no improvement in the loss on validation data over 100 epochs.

To evaluate HVAE after training is complete, we estimate out-of-sample negative log likelihoods (NLLs) using 1000 importance samples from the HVAE approximate posterior.
For each trained model, we estimate NLL three times, noting that the standard deviation over these three estimates is no larger than 0.12 nats.
We report the average NLL values over either two or three different initializations (in addition to the three NLL estimates for each trained model) for several choices of tempering and leapfrog steps in \autoref{tab:mnist_nll}.
A full accounting of the tests is given in the supplementary material.
We also consider an HVAE scheme in which we allow $\varepsilon$ to vary across layers of the flow and report the results.

From \autoref{tab:mnist_nll}, we notice that generally increasing the inhomogeneity in the dynamics improves the test NLL values.
For example, free tempering is the most successful tempering scheme, and varying the leapfrog step size $\varepsilon$ across layers also improves results.
We also notice that increasing the number of leapfrog steps does not always improve the performance, as $K = 15$ provides the best results in free tempering schemes.
We believe that the improvement in HVAE over the base VAE scheme can be attributed to a more expressive approximate posterior, as we can see that samples from the HVAE approximate posterior exhibit non-negligible covariance across dimensions.
As in \citet{salimans2015markov}, we are also able to improve upon the base model by adding our time-inhomogeneous Hamiltonian dynamics on top, but in a simplified regime without referring to learned reverse kernels.
\citet{rezende2015variational} report only lower bounds on the log-likelihood for NFs, which are indeed lower than our log-likelihood estimates, although they use a much larger number of variational parameters.

\begin{table}
	\caption{Estimated NLL values for HVAE on MNIST. The base VAE achieves an NLL of 83.20. A more detailed version of this table is included in the supplementary material.}
	\label{tab:mnist_nll}
	\ra{1.2}
	\centering
	\begin{tabular}{@{}rrrrcrrr@{}}
		\toprule
		& \multicolumn{3}{c}{$\varepsilon$ fixed across layers}
		& \phantom{abc}
		& \multicolumn{3}{c}{$\varepsilon$ varied across layers} \\
		\cmidrule{2-4} \cmidrule{6-8}
		& $T =$ Free & $T =$ Fixed & $T= $None && $T =$ Free & $T =$ Fixed & $T= $None \\
						\midrule
		$K=1$  	& N/A   & 83.32 & 83.17 && N/A   & N/A   & N/A   \\
		$K=5$  	& 83.09 & 83.26 & 83.68 && 83.01 & 82.94 & 83.35 \\
		$K=10$ 	& 82.97 & 83.26 & 83.40 && 82.62 & 82.87 & 83.25 \\
		$K=15$ 	& 82.78 & 83.56 & 83.82 && 82.62 & 83.09 & 82.94 \\
		$K=20$ 	& 82.93 & 83.18 & 83.33 && 82.83 & 82.85 & 82.93 \\
		\bottomrule
	\end{tabular}
\end{table}

\section{Conclusion and Discussion}
We have proposed a principled way to exploit Hamiltonian dynamics within stochastic variational inference.
Contrary to previous methods \cite{salimans2015markov,wolf2016variational}, our algorithm does not rely on learned reverse Markov kernels and benefits from the use of tempering ideas.
Additionally, we can use the reparameterization trick to obtain unbiased estimators of gradients of the ELBO.
 The resulting HVAE can be interpreted as a target-driven normalizing flow which requires the evaluation of a few gradients of the log-likelihood associated to a single data point at each stochastic gradient step.
 However, the Jacobian computations required for the ELBO are trivial.
In our experiments, the robustness brought about by the use of target-informed dynamics can reduce the number of parameters that must be trained and improve generalizability.

We note that, although we have fewer parameters to optimize, the memory cost of using HVAE and target-informed dynamics could become prohibitively large if the memory required to store evaluations of $\nabla_z \log p_\theta(x, z)$ is already extremely large.
 Evaluating these gradients is not a requirement of VAEs or standard normalizing flows.
 However, we have shown that in the case of a fairly large generative network we are still able to evaluate gradients and backpropagate through the layers of the flow.
 Further tests explicitly comparing HVAE with VAEs and normalizing flows in various memory regimes are required to determine in what cases one method should be used over the other.

There are numerous possible extensions of this work. Hamiltonian dynamics preserves the Hamiltonian and hence also the corresponding target distribution, but there exist other deterministic dynamics which leave the target distribution invariant but not the Hamiltonian. This includes the {N}os{\'e}-{H}oover thermostat.
It is possible to directly use these dynamics instead of the Hamiltonian dynamics within the framework developed in \autoref{sec:deterministic}. In continuous-time, related ideas have appeared in physics \cite{cuendet2006statistical,procacci2006crooks,scholl2006proof}. This comes at the cost of more complicated Jacobian calculations. 
The ideas presented here could also be coupled with the methodology proposed in \cite{heng2017controlled} -- we conjecture that this could reduce the variance of the estimator \eqref{eq:generalizedJarzynski} by an order of magnitude.

\subsubsection*{Acknowledgments}
Anthony L.\ Caterini is a Commonwealth Scholar, funded by the UK government.

\bibliography{refs}

\begin{thebibliography}{28}
\providecommand{\natexlab}[1]{#1}
\providecommand{\url}[1]{\texttt{#1}}
\expandafter\ifx\csname urlstyle\endcsname\relax
  \providecommand{\doi}[1]{doi: #1}\else
  \providecommand{\doi}{doi: \begingroup \urlstyle{rm}\Url}\fi

\bibitem[Abadi et~al.(2015)]{tensorflow2015-whitepaper}
Mart\'{\i}n Abadi et~al.
\newblock {TensorFlow}: Large-scale machine learning on heterogeneous systems,
  2015.
\newblock URL \url{https://www.tensorflow.org/}.
\newblock Software available from tensorflow.org.

\bibitem[Berg et~al.(2018)Berg, Hasenclever, Tomczak, and
  Welling]{berg2018sylvester}
Rianne van~den Berg, Leonard Hasenclever, Jakub~M Tomczak, and Max Welling.
\newblock Sylvester normalizing flows for variational inference.
\newblock \emph{arXiv preprint arXiv:1803.05649}, 2018.

\bibitem[Burda et~al.(2016)Burda, Grosse, and
  Salakhutdinov]{burda2015importance}
Yuri Burda, Roger Grosse, and Ruslan Salakhutdinov.
\newblock Importance weighted autoencoders.
\newblock In \emph{The 4th International Conference on Learning Representations
  (ICLR)}, 2016.

\bibitem[Crooks(1998)]{crooks1998nonequilibrium}
Gavin~E Crooks.
\newblock Nonequilibrium measurements of free energy differences for
  microscopically reversible {M}arkovian systems.
\newblock \emph{Journal of Statistical Physics}, 90\penalty0 (5-6):\penalty0
  1481--1487, 1998.

\bibitem[Cuendet(2006)]{cuendet2006statistical}
Michel~A Cuendet.
\newblock Statistical mechanical derivation of {J}arzynski’s identity for
  thermostated non-hamiltonian dynamics.
\newblock \emph{Physical Review Letters}, 96\penalty0 (12):\penalty0 120602,
  2006.

\bibitem[Del~Moral et~al.(2006)Del~Moral, Doucet, and Jasra]{del2006sequential}
Pierre Del~Moral, Arnaud Doucet, and Ajay Jasra.
\newblock Sequential {M}onte {C}arlo samplers.
\newblock \emph{Journal of the Royal Statistical Society: Series B (Statistical
  Methodology)}, 68\penalty0 (3):\penalty0 411--436, 2006.

\bibitem[Dosovitskiy et~al.(2015)Dosovitskiy, Springenberg, and
  Brox]{dosovitskiy2015learning}
Alexey Dosovitskiy, Jost~Tobias Springenberg, and Thomas Brox.
\newblock Learning to generate chairs with convolutional neural networks.
\newblock In \emph{Computer Vision and Pattern Recognition (CVPR), 2015 IEEE
  Conference on}, pages 1538--1546. IEEE, 2015.

\bibitem[Glasserman(1991)]{glasserman1991gradient}
Paul Glasserman.
\newblock \emph{Gradient estimation via perturbation analysis}, volume 116.
\newblock Springer Science \& Business Media, 1991.

\bibitem[Heng et~al.(2017)Heng, Bishop, Deligiannidis, and
  Doucet]{heng2017controlled}
Jeremy Heng, Adrian~N Bishop, George Deligiannidis, and Arnaud Doucet.
\newblock Controlled sequential {M}onte {C}arlo.
\newblock \emph{arXiv preprint arXiv:1708.08396}, 2017.

\bibitem[Hoffman(2017)]{hoffman2017learning}
Matthew~D Hoffman.
\newblock Learning deep latent {G}aussian models with {M}arkov chain {M}onte
  {C}arlo.
\newblock In \emph{International Conference on Machine Learning}, pages
  1510--1519, 2017.

\bibitem[Hoffman and Gelman(2014)]{hoffman2014no}
Matthew~D Hoffman and Andrew Gelman.
\newblock The no-u-turn sampler: adaptively setting path lengths in hamiltonian
  monte carlo.
\newblock \emph{Journal of Machine Learning Research}, 15\penalty0
  (1):\penalty0 1593--1623, 2014.

\bibitem[Jarzynski(1997)]{jarzynski1997nonequilibrium}
Christopher Jarzynski.
\newblock Nonequilibrium equality for free energy differences.
\newblock \emph{Physical Review Letters}, 78\penalty0 (14):\penalty0 2690,
  1997.

\bibitem[Jarzynski(2000)]{jarzynski2000hamiltonian}
Christopher Jarzynski.
\newblock Hamiltonian derivation of a detailed fluctuation theorem.
\newblock \emph{Journal of Statistical Physics}, 98\penalty0 (1-2):\penalty0
  77--102, 2000.

\bibitem[Kingma and Ba(2014)]{kingma2014adam}
Diederik~P Kingma and Jimmy Ba.
\newblock Adam: A method for stochastic optimization.
\newblock \emph{arXiv preprint arXiv:1412.6980}, 2014.

\bibitem[Kingma and Welling(2014)]{kingma2013auto}
Diederik~P Kingma and Max Welling.
\newblock Auto-encoding variational {B}ayes.
\newblock In \emph{The 2nd International Conference on Learning Representations
  (ICLR)}, 2014.

\bibitem[Kingma et~al.(2016)Kingma, Salimans, Jozefowicz, Chen, Sutskever, and
  Welling]{kingma2016improved}
Diederik~P Kingma, Tim Salimans, Rafal Jozefowicz, Xi~Chen, Ilya Sutskever, and
  Max Welling.
\newblock Improved variational inference with inverse autoregressive flow.
\newblock In \emph{Advances in Neural Information Processing Systems}, pages
  4743--4751, 2016.

\bibitem[Maddison et~al.(2017)Maddison, Lawson, Tucker, Heess, Norouzi, Mnih,
  Doucet, and Teh]{maddison2017filtering}
Chris~J Maddison, John Lawson, George Tucker, Nicolas Heess, Mohammad Norouzi,
  Andriy Mnih, Arnaud Doucet, and Yee Teh.
\newblock Filtering variational objectives.
\newblock In \emph{Advances in Neural Information Processing Systems}, pages
  6576--6586, 2017.

\bibitem[Neal(2001)]{neal2001annealed}
Radford~M Neal.
\newblock Annealed importance sampling.
\newblock \emph{Statistics and Computing}, 11\penalty0 (2):\penalty0 125--139,
  2001.

\bibitem[Neal(2005)]{neal2005hamiltonian}
Radford~M Neal.
\newblock Hamiltonian importance sampling.
\newblock \url{www.cs.toronto.edu/pub/radford/his-talk.ps}, 2005.
\newblock Talk presented at the Banff International Research Station (BIRS)
  workshop on Mathematical Issues in Molecular Dynamics.

\bibitem[Neal et~al.(2011)]{neal2011mcmc}
Radford~M Neal et~al.
\newblock {MCMC} using {H}amiltonian dynamics.
\newblock \emph{Handbook of Markov Chain Monte Carlo}, 2\penalty0 (11), 2011.

\bibitem[Procacci et~al.(2006)Procacci, Marsili, Barducci, Signorini, and
  Chelli]{procacci2006crooks}
Piero Procacci, Simone Marsili, Alessandro Barducci, Giorgio~F Signorini, and
  Riccardo Chelli.
\newblock Crooks equation for steered molecular dynamics using a
  {N}os{\'e}-{H}oover thermostat.
\newblock \emph{The Journal of Chemical Physics}, 125\penalty0 (16):\penalty0
  164101, 2006.

\bibitem[Rezende and Mohamed(2015)]{rezende2015variational}
Danilo Rezende and Shakir Mohamed.
\newblock Variational inference with normalizing flows.
\newblock In \emph{International Conference on Machine Learning}, pages
  1530--1538, 2015.

\bibitem[Rezende et~al.(2014)Rezende, Mohamed, and
  Wierstra]{rezende2014stochastic}
Danilo Rezende, Shakir Mohamed, and Daan Wierstra.
\newblock Stochastic backpropagation and approximate inference in deep
  generative models.
\newblock In \emph{International Conference on Machine Learning}, pages
  1278--1286, 2014.

\bibitem[Salakhutdinov and Murray(2008)]{salakhutdinov2008quantitative}
Ruslan Salakhutdinov and Iain Murray.
\newblock On the quantitative analysis of deep belief networks.
\newblock In \emph{Proceedings of the 25th international conference on Machine
  learning}, pages 872--879. ACM, 2008.

\bibitem[Salimans et~al.(2015)Salimans, Kingma, and
  Welling]{salimans2015markov}
Tim Salimans, Diederik~P Kingma, and Max Welling.
\newblock Markov chain {M}onte {C}arlo and variational inference: Bridging the
  gap.
\newblock In \emph{International Conference on Machine Learning}, pages
  1218--1226, 2015.

\bibitem[Sch{\"o}ll-Paschinger and Dellago(2006)]{scholl2006proof}
E~Sch{\"o}ll-Paschinger and Christoph Dellago.
\newblock A proof of {J}arzynski’s nonequilibrium work theorem for dynamical
  systems that conserve the canonical distribution.
\newblock \emph{The Journal of Chemical Physics}, 125\penalty0 (5):\penalty0
  054105, 2006.

\bibitem[Tieleman and Hinton(2012)]{tieleman2012lecture}
Tijmen Tieleman and Geoffrey Hinton.
\newblock Lecture 6.5-rmsprop: Divide the gradient by a running average of its
  recent magnitude.
\newblock \emph{COURSERA: Neural networks for machine learning}, 4\penalty0
  (2):\penalty0 26--31, 2012.

\bibitem[Wolf et~al.(2016)Wolf, Karl, and van~der Smagt]{wolf2016variational}
Christopher Wolf, Maximilian Karl, and Patrick van~der Smagt.
\newblock Variational inference with {H}amiltonian {M}onte {C}arlo.
\newblock \emph{arXiv preprint arXiv:1609.08203}, 2016.

\end{thebibliography}

\newpage
\appendix
\appendixpage

\section{Full List of Tests on MNIST}

\autoref{tab:fixed_eps} and \autoref{tab:varied_eps} display the list of test runs of HVAE on MNIST. The number of flow steps is denoted by $K$. The total number of epochs varies in training because of early stopping. The ELBO and NLL estimates are generated using 1000 importance samples from the HVAE approximate posterior; this procedure is run 3 times for each seed. The average and standard deviation of these estimates over the three runs is displayed. \autoref{tab:fixed_eps} refers to HVAE tests in which $\varepsilon$ was fixed across flow layers, whereas \autoref{tab:varied_eps} refers to HVAE tests in which $\varepsilon$ was allowed to vary across flow layers.

\begin{table}[H]
\caption{List of tests of HVAE for $\varepsilon$ fixed across flow layers}
\label{tab:fixed_eps}
\ra{1.2}
\centering
\begin{tabular}{rrrrrr}
	\toprule
	\bfseries K & \bfseries Tempering & \bfseries Seed & \bfseries Total Epochs & \bfseries ELBO Estimate & \bfseries NLL Estimate
	\begin{filecontents*}{mnist_fixed_eps.csv}
Model,K,Tempering,Seed,TotalEpochs,ELBO,ELBOError,NLL,NLLError
hvae,1,fixed,85547,1158,86.49,0.05,83.30,0.06
hvae,1,fixed,12345,1131,86.53,0.02,83.33,0.02
hvae,1,none,85547,1374,86.43,0.08,83.29,0.07
hvae,1,none,12345,1519,86.21,0.04,83.04,0.04
hvae,5,fixed,85547,998,86.69,0.01,83.36,0.02
hvae,5,fixed,12345,1070,86.40,0.04,83.16,0.04
hvae,5,free,85547,1046,86.45,0.05,83.16,0.04
hvae,5,free,12345,1141,86.31,0.11,83.02,0.10
hvae,5,none,85547,975,86.82,0.04,83.68,0.05
hvae,5,none,12345,959,86.82,0.02,83.68,0.02
hvae,10,fixed,85547,991,86.75,0.03,83.50,0.03
hvae,10,fixed,12345,1340,86.26,0.05,83.02,0.04
hvae,10,free,85547,1404,86.06,0.05,82.73,0.05
hvae,10,free,12345,819,86.65,0.04,83.18,0.03
hvae,10,free,98765,846,86.47,0.08,82.99,0.08
hvae,10,none,85547,1425,86.47,0.07,83.24,0.06
hvae,10,none,12345,1035,86.78,0.02,83.56,0.03
hvae,15,fixed,85547,666,87.11,0.02,83.93,0.02
hvae,15,fixed,12345,1204,86.51,0.08,83.19,0.07
hvae,15,free,85547,1303,86.19,0.03,82.75,0.04
hvae,15,free,12345,1028,86.16,0.06,82.81,0.06
hvae,15,none,85547,785,87.05,0.06,83.87,0.06
hvae,15,none,12345,989,86.95,0.03,83.76,0.03
hvae,20,fixed,85547,1103,86.30,0.08,83.18,0.07
hvae,20,fixed,12345,1126,86.32,0.09,83.18,0.09
hvae,20,free,85547,1028,86.28,0.03,82.96,0.03
hvae,20,free,12345,1131,86.27,0.02,82.89,0.02
hvae,20,none,85547,1318,86.36,0.06,83.30,0.06
hvae,20,none,12345,1299,86.44,0.05,83.35,0.05
\end{filecontents*}
\csvreader[head to column names]{mnist_fixed_eps.csv}{}
	{\\\hline \K & \Tempering & \Seed & \TotalEpochs & \ELBO \hspace{0.5em} (\ELBOError) & \NLL \hspace{0.5em} (\NLLError)} \\
	\bottomrule
\end{tabular}
\end{table}

\begin{table}[H]
\caption{List of tests of HVAE for $\varepsilon$ fixed across flow layers}
\label{tab:varied_eps}
\ra{1.2}
\centering
\begin{tabular}{rrrrrr}
	\toprule
	\bfseries K & \bfseries Tempering & \bfseries Seed & \bfseries Total Epochs & \bfseries ELBO Estimate & \bfseries NLL Estimate
	\begin{filecontents*}{mnist_varied_eps.csv}
Model,K,Tempering,Seed,TotalEpochs,ELBO,ELBOError,NLL,NLLError
hvae,5,fixed,85547,1229,86.28,0.04,83.10,0.04
hvae,5,fixed,12345,1579,85.89,0.02,82.77,0.02
hvae,5,free,85547,1116,86.24,0.02,83.11,0.02
hvae,5,free,12345,1302,86.07,0.07,82.91,0.07
hvae,5,none,85547,1014,86.50,0.06,83.38,0.06
hvae,5,none,12345,959,86.46,0.01,83.32,0.01
hvae,10,fixed,85547,1473,86.01,0.05,82.88,0.05
hvae,10,fixed,12345,1508,86.01,0.01,82.85,0.01
hvae,10,free,85547,1303,85.98,0.03,82.69,0.03
hvae,10,free,12345,1680,85.61,0.02,82.37,0.02
hvae,10,free,98765,846,86.17,0.08,82.80,0.09
hvae,10,none,85547,1035,86.44,0.08,83.35,0.07
hvae,10,none,12345,1276,86.29,0.12,83.14,0.12
hvae,15,fixed,85547,998,86.41,0.01,83.20,0.02
hvae,15,fixed,12345,1070,86.21,0.02,82.97,0.02
hvae,15,free,85547,1303,85.92,0.03,82.57,0.02
hvae,15,free,12345,1131,85.96,0.03,82.66,0.03
hvae,15,none,85547,1425,86.08,0.06,82.96,0.05
hvae,15,none,12345,1355,86.06,0.01,82.92,0.01
hvae,20,fixed,85547,1473,85.85,0.05,82.71,0.06
hvae,20,fixed,12345,1200,86.18,0.06,82.98,0.05
hvae,20,free,85547,1404,86.02,0.05,82.75,0.05
hvae,20,free,12345,1131,86.09,0.03,82.90,0.02
hvae,20,none,85547,1596,85.97,0.03,82.83,0.02
hvae,20,none,12345,1327,86.17,0.06,83.03,0.06
\end{filecontents*}
\csvreader[head to column names]{mnist_varied_eps.csv}{}
	{\\\hline \K & \Tempering & \Seed & \TotalEpochs & \ELBO \hspace{0.5em} (\ELBOError) & \NLL \hspace{0.5em} (\NLLError)} \\
	\bottomrule	
\end{tabular}
\end{table}

\autoref{tab:vae} displays the list of test runs of the base VAE on MNIST. ELBO and NLL estimates are again generated by importance sampling, but this time from the learned VAE approximate posterior.

\begin{table}[H]
\caption{List of tests of VAE}
\label{tab:vae}
\ra{1.2}
\centering
\begin{tabular}{rrrr}
	\toprule
	\bfseries Seed & \bfseries Total Epochs & \bfseries ELBO Estimate & \bfseries NLL Estimate
	\begin{filecontents*}{mnist_cnn.csv}
Model,Seed,TotalEpochs,ELBO,ELBOError,NLL,NLLError
cnn,85547,1524,86.41,0.01,83.25,0.01
cnn,98765,1301,86.42,0.07,83.19,0.07
cnn,12345,1381,86.43,0.07,83.15,0.07
\end{filecontents*}
\csvreader[head to column names]{mnist_cnn.csv}{}
	{\\\hline \Seed & \TotalEpochs & \ELBO \hspace{0.5em} (\ELBOError) & \NLL \hspace{0.5em} (\NLLError)} \\
	\bottomrule	
\end{tabular}
\end{table}

\end{document}